\documentclass{article}
\usepackage{spconf,amsmath,graphicx,hyperref}
\usepackage[table]{xcolor}
\usepackage{amssymb}
\usepackage[table,xcdraw]{xcolor} 
\usepackage{colortbl} 

\title{Causal Fingerprints of AI Generative Models}
\name{Hui Xu$^1$, Chi Liu$^{1,2,3*}$\thanks{* Corresponding Author: Chi Liu (chiliu@cityu.edu.mo)}, Congcong Zhu$^1$, Minghao Wang$^1$, Youyang Qu$^4$, Longxiang Gao$^4$}
\address{$^1$Faculty of Data Science,City University of Macau, Macao SAR, China\\ 
$^2$Nanjing University of Science and Technology, Nanjing, China\\ 
$^3$Zhejiang Cloudpeng Technology Co., Ltd., China\\ 
$^4$Key Laboratory of Computing Power Network and Information Security, \\Ministry of Education, Qilu University of Technology (Shandong Academy of Sciences)}

\begin{document}
\topmargin=0mm
\ninept
\maketitle
\begin{abstract}

AI generative models leave implicit traces in their generated images, which are commonly referred to as model fingerprints and are exploited for source attribution. Prior methods rely on model-specific cues or synthesis artifacts, yielding limited fingerprints that may generalize poorly across different generative models. We argue that a complete model fingerprint should reflect the causality between image provenance and model traces, a direction largely unexplored. To this end, we conceptualize the \emph{causal fingerprint} of generative models, and propose a causality-decoupling framework that disentangles it from image-specific content and style in a semantic-invariant latent space derived from pre-trained diffusion reconstruction residual. We further enhance fingerprint granularity with diverse feature representations. We validate causality by assessing attribution performance across representative GANs and diffusion models and by achieving source anonymization using counterfactual examples generated from causal fingerprints. Experiments show our approach outperforms existing methods in model attribution, indicating strong potential for forgery detection, model copyright tracing, and identity protection.

\end{abstract}

\section{Introduction}
\label{sec:intro}
The rapid evolution of generative models has significantly improved AI-generated content (AIGC), particularly in producing highly realistic images. However, this creates challenges for model attribution, which aims to identify the correct source model that generated an image. Model attribution is crucial for AIGC safety \cite{C53,C55}: it offers an auditing mechanism of image authenticity to counter malicious forgeries \cite{C13,C14}; meanwhile, it makes the source model traceable in its outputs, thereby safeguarding the model owner's copyright from pirates\cite{C15,C16,C25}.

The current main approaches to model attribution include forgery detection, watermarking, and fingerprinting. Forgery detection formulates a binary classification to distinguish real from AI-generated images; it is a passive method and fails to support fine-grained source attribution\cite{C1,C2,C3,C4,C5,C6}. Watermarking proactively embeds verifiable source information into the image; but it requires modifying the original generative model and can corrupt the original image. In contrast, fingerprinting exposes and analyzes inherent traces left by the model in the image, enabling explicit model identification without altering the original model and images, and is therefore more efficient and practical\cite{C8,C9,C21,C22,C7}. 


Prior work typically defines model fingerprints as residual noise or image artifacts introduced by the generator and adopts a feature-extraction paradigm. While effective for identifying specific models, these methods depend heavily on model-specific cues and generalize poorly to models that differ in these predefined features. For instance, some Generative Adversarial Networks (GANs) produce checkerboard artifacts that have been exploited as fingerprints in prior studies \cite{C21,C22,C54}; however, this cue does not transfer to models that exhibit no checkerboard patterns, such as some diffusion models. We argue that such feature-based approaches yield fragmentary and limited fingerprints, while the comprehensive fingerprint that captures true causality between image provenance and model traces, what we term as the \emph{causal fingerprint}, remains largely overlooked. 

\begin{figure}[htbp!]
  \centering
  \includegraphics[width=0.43\linewidth]{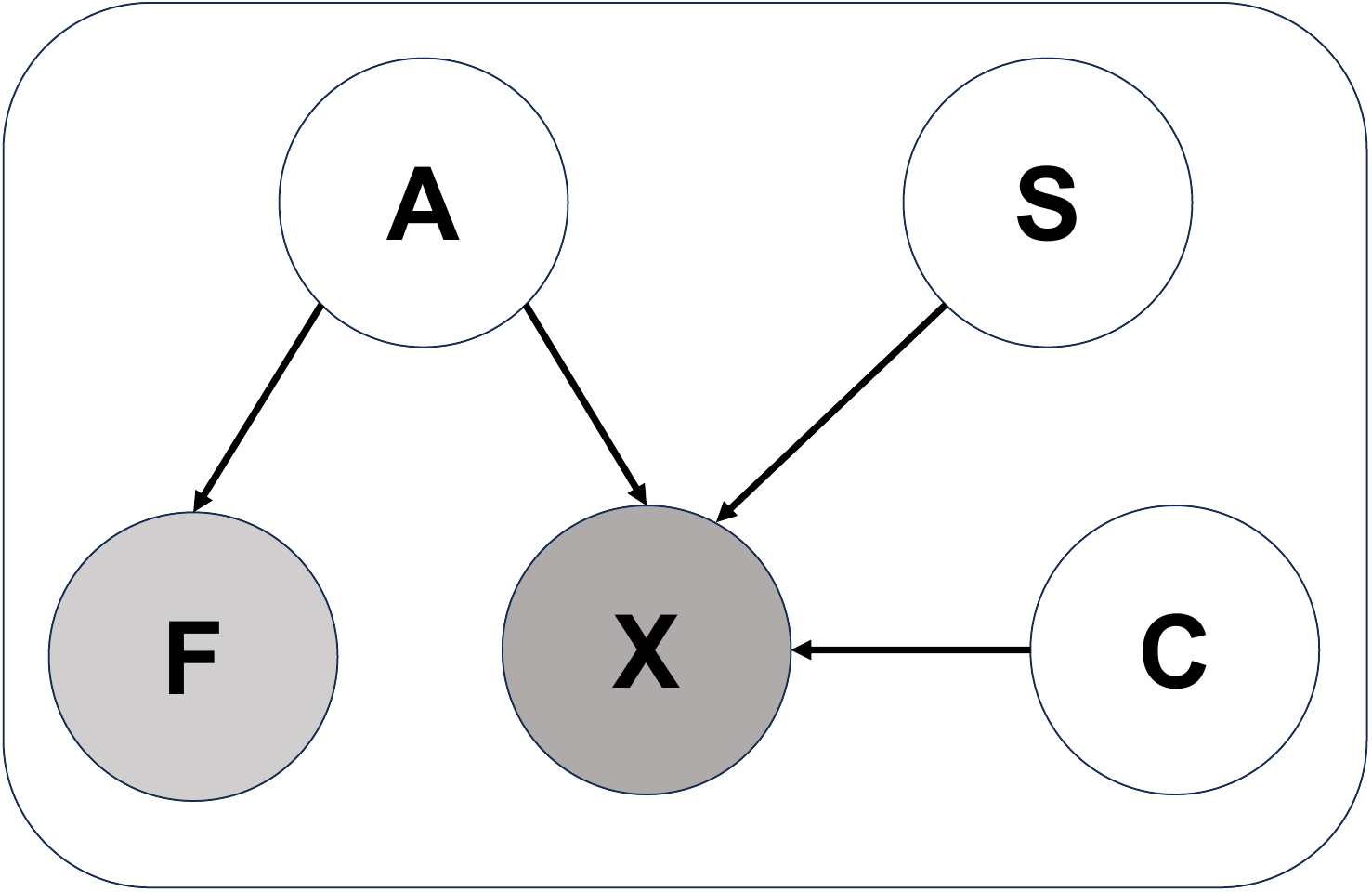}
\caption{The causal relationship between model fingerprints and AI-generated images, where $X$, $F$, $A$, $C$, and $S$ represent the image, the model fingerprint of source generation model, the image's artifacts, content, and style, respectively.}
\label{fig:res1}
\end{figure}

To close this gap, we present the first definition of causal fingerprint (CF) of AI generative models, together with a causality-inspired disentanglement framework that separates CFs from image-specific content and style. The framework extracts CFs within a semantics-invariant latent space derived from pretrained diffusion reconstruction residuals \cite{C6}, and further increases fingerprint granularity by exploring and embedding broader, more diverse feature representations \cite{C10,C55,C56}. Moreover, the causal properties of the extracted fingerprints enable the construction of counterfactual fingerprints that facilitate image source obfuscation, based on which we design a model anonymization method using fingerprint-constrained PGD adversarial perturbations. To assess the causality of the fingerprints and their utility for model attribution, we conduct experiments on a challenging AI-generated image benchmark and compare against six representative baselines from different categories for attributing four generative models, including GANs and diffusion models. Additional analyses, including an ablation study on feature representations for fine-grained fingerprints, fingerprint visualizations, and model anonymization using counterfactual fingerprints, are performed to further validate the presence of fingerprints and their causality. Experimental results demonstrate that our decoupling framework successfully extracts CFs, surpasses prior methods in model attribution, and achieves model anonymization via counterfactual examples, indicating strong potential for forgery detection, model copyright tracing, and identity protection.

\begin{figure*}[htb]

  \centering
  \includegraphics[width=0.88\textwidth]{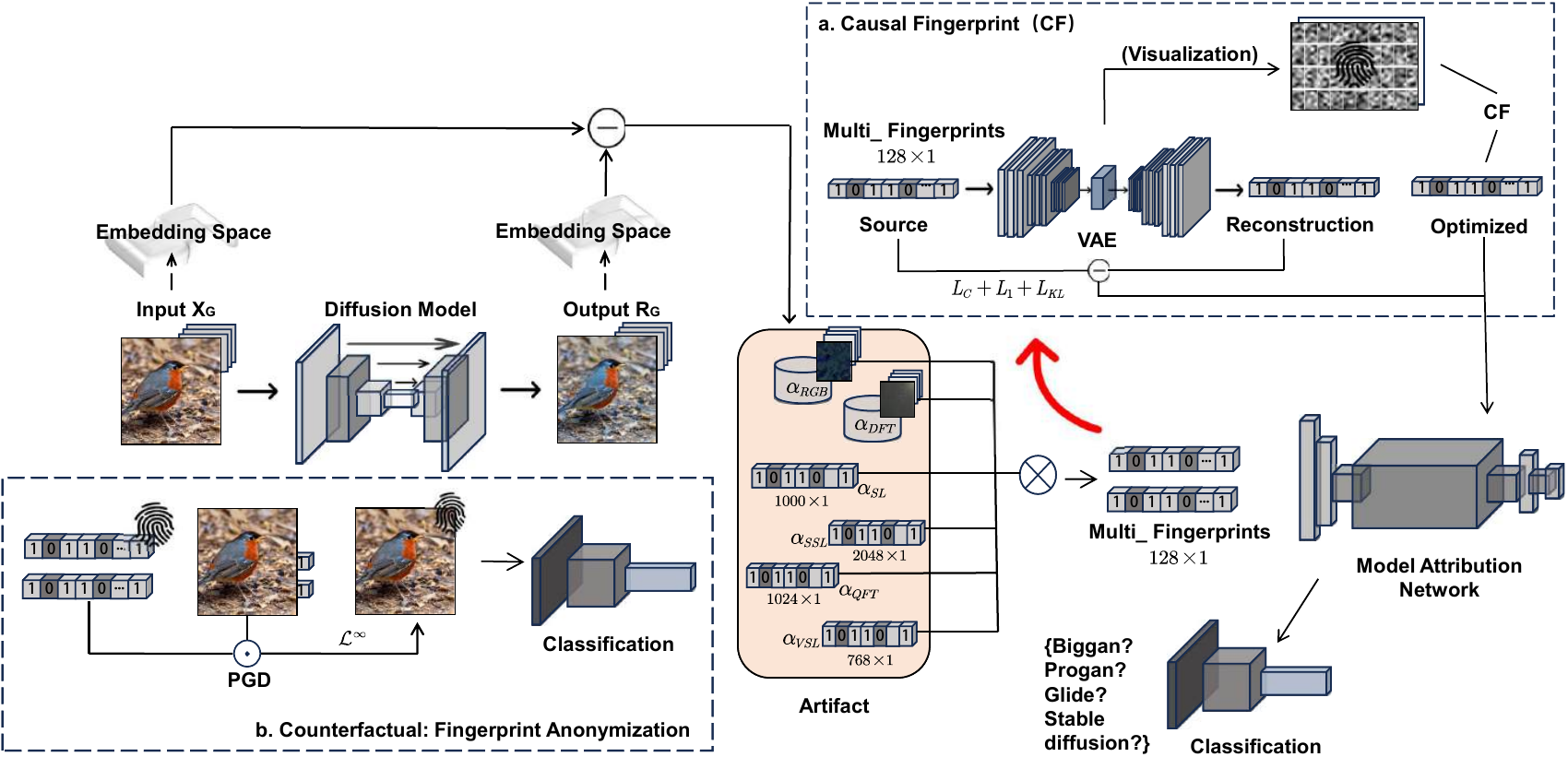}
%
\caption{Architecture diagram of our causal fingerprint decoupling scheme based on the semantic-invariant  space. We separate causal fingerprints by end-to-end training a variational autoencoder, then train a classifier using a model attribution network. b is the framework diagram of the counterfactual method.}
\label{fig:res2}
\end{figure*}

\section{Method}
\label{sec:pagestyle}

\subsection{Definition of Causal Fingerprint (CF)}
\label{ssec:2.1}
Model fingerprints are features within images generated by AIGC models, related to model architecture and algorithmic configuration. They reflect causal representations within the generation process, stemming from its non-random nature. A generated image $X$ comprises content $C$, style $S$, and artefacts $A$. The fingerprint $F$ is directly determined solely by the artefacts $A$. Using causal graph modelling(Fig. \ref{fig:res1}), $C$, $S$, and $A$ are direct causes of $X$, while $F$ has $A$ as its direct cause, satisfying $C\perp \!\!\! \perp S\perp \!\!\! \perp A$. However, given $X=x$, $C\not \! \perp \!\!\! \perp S\not \! \perp \!\!\! \perp A\mid X$, leading to spurious correlations among $C$, $S$, $A$, and $F$, since $A$ is the direct cause of $F$ \cite{C42}. Mathematically, the causality of the fingerprint $F$ is defined as:
\begin{equation}
F = f(A),
\label{eq:1}
\end{equation}
\begin{equation}
F_G = \sum_{s \in \mathcal{E}} w_s \phi_s(r_{dire}),
\label{eq:2}
\end{equation}
Here, $f(A)$ denotes the mapping function that extracts fingerprints from artifact $A$, wherein artifact $A$ captures specific traces during the model training process, independent of content $C$ and style $S$. To further enrich the fingerprint representation, causal fingerprints are defined by weighting and concatenating the differences in projections across multiple semantically invariant embedding spaces, as shown in Equation \ref{eq:2}, where $\mathcal{E}$ denotes the set of semantically invariant embedding spaces, $\phi_s(\cdot)$ represents the feature extraction function for the $s$th embedding space, $w_s$ denotes the weighting coefficient, and $r_{dire}$ denotes the residual computed via the pre-trained Diffusion-based Residual Model (DIRE), detailed in Sec. \ref{ssec:2.2}.

\subsection{Decouple CF in Semantic-Invariant Latent Space}
\label{ssec:2.2}
To achieve causal disentanglement, the model fingerprint $F$ must be decoupled from the image content $C$ and style $S$, ensuring the fingerprint reflects the model's intrinsic properties rather than the specific semantic information of the generated image. Let $\mathcal{E}$ denote a semantic-invariant latent space (SILS). The generative model $G$ maps latent codes $z$ to images $X = G(z)$. The projection difference is defined as the divergence between embeddings of images generated by different models within $\mathcal{E}$, after eliminating semantically related components, namely:
\begin{equation}
\Delta_{\mathcal{E}} = \phi_{\mathcal{E}}(X_1) - \phi_{\mathcal{E}}(X_2),
\label{eq:3}
\end{equation}
\begin{equation}
\hat{X} = \mathcal{E}(X), \quad r = X - \hat{X},
\label{eq:4}
\end{equation}
where $\phi_{\mathcal{E}}(\cdot)$ denotes the feature extraction function within the embedding space $\mathcal{E}$, and $X_1$ and $X_2$ represent images generated by distinct models. The projected differences capture only factors causally related to the generation process—such as inductive biases inherent in the model architecture or artefacts arising during optimisation—rather than semantic content.

To compute the projection difference in the semantically invariant embedding space, we propose employing the reconstruction residual method. The reconstruction residual is defined as the difference between the generated image $X$ and its corresponding image $\hat{X}$ reconstructed via a specific generative model, as shown in Equation \ref{eq:4}.One of them is $\hat{X}$, the image reconstructed with high fidelity at a 1:1 ratio through the embedding space $\mathcal{E}$. The residual $r$ captures specific artefacts associated with the model's generation process, independent of the image's semantic information. To enhance the robustness of causal fingerprint extraction, the pre-trained Diffusion Reconstruction Residual (DIRE) \cite{C6} model is selected for its high semantic richness and ability to preserve structural details during image reconstruction. It maintains semantic consistency in the reconstructed image $\hat{X}$, ensuring the residual $r$ captures only model-specific artefacts. Its pre-trained nature endows it with generalisation capability across diverse data distributions, effectively mitigating semantic correlations and domain bias. Consequently, the residual $r_{dire}$ generated by DIRE is expressed as:
\begin{equation}
r_{dire} = X - \hat{X}_{dire},
\label{eq:5}
\end{equation}
where $\hat{X}_{dire}$ represents the image reconstructed by the DIRE model, which is used for calculating the causal fingerprint $F_G$ in Sec. \ref{ssec:2.1}.
\subsection{Expanding SILS for Fine-gained CF}
\label{ssec:2.3}
\begin{figure}[!htb]
  \centering
  \includegraphics[width=0.93\linewidth]{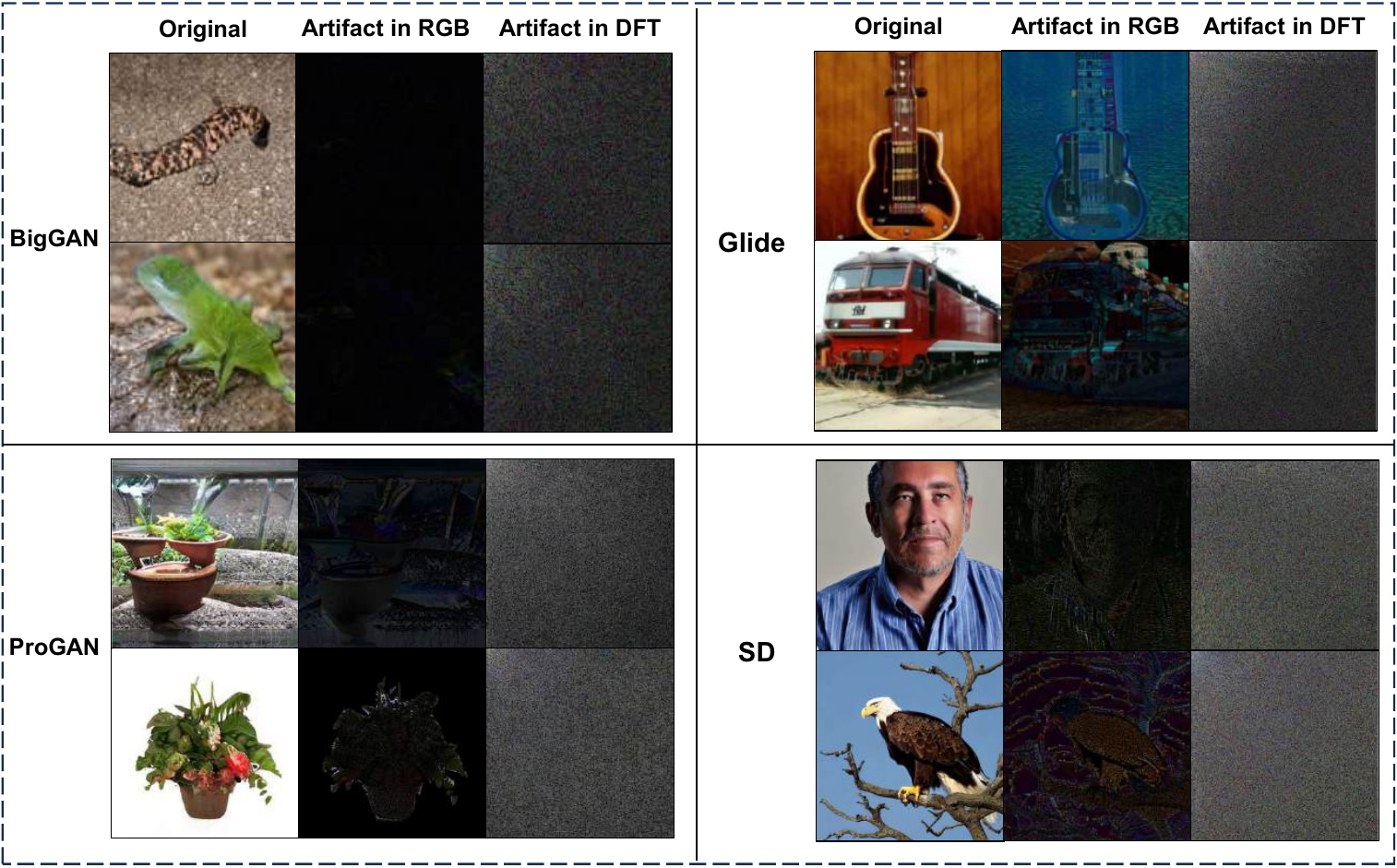}
\caption{Examples of artefacts in extracted RGB and DFT spaces. Whereas QFT, SL, SSL and VSL, being vectors of lengths $1024$, $1000$, $2048$ and $768$ respectively (within the embedding spaces of their respective pre-trained network architectures).}
\label{fig:res3}
\end{figure}

To realise the causal fingerprint $F_G$ defined in Sec.\ref{ssec:2.1}, we consider six semantically invariant latent spaces for extracting image artefacts, thereby enriching the granularity representation of the fingerprint. Fig. \ref{fig:res3} illustrates artifacts in both the RGB space and the frequency domain. For the RGB space, we utilise pixel-valued RGB images; for the frequency space, we convert RGB images into 2D spectra by applying a Discrete Cosine Transform (DCT) to each channel. Additionally, we convert RGB images into grayscale, apply a Fast Fourier Transform to these grayscale images to generate 2D spectra, and then extract the low-frequency components (QFT). For the embedding spaces of supervised learning methods (SL and VSL), we respectively employed the encoder head of ResNet101 \cite{C46} pre-trained on ImageNet and extracted class-token features using a pre-trained ViT model \cite{C45}; for the embedding space of the self-supervised learning method (SSL), we utilised the encoder head of pre-trained DINO ResNet50 \cite{C47}. By weighting and fusing the projection differences across these embedding spaces, the generated causal fingerprint $F_G$ comprehensively captures model-specific artefacts from the generation process, enhancing both the robustness and attribution accuracy of fingerprint extraction.

\subsection{Network Architecture}
\label{ssec:2.4}
\textbf{Fingerprint Visualization.}
\label{ssec:2.4.1}
Since the model's causal fingerprint is a 128-dimensional vector, we employ a variational autoencoder (VAE) to achieve optimization and visual decoupling \cite{C8}. The encoder compresses the 128-dimensional features into a 64-dimensional latent representation, while the decoder maps back to the 128-dimensional fingerprint feature space to generate reconstructed features. By reparameterizing the sampling \( z \sim \mathcal{N}(\mu, \sigma^2) \), we optimize the causal attribution loss function using \(L_c\) loss, \(L_1\) loss, and \(L_{KL}\) divergence. The trained latent vector \( z \) is reshaped into an \( 8 \times 8 \) matrix for visualizing causal fingerprints. 
\begin{equation}
L_c=\frac{1}{N}\sum_{i=1}^N{(}x_i-\hat{x}_i)^2,
\label{eq:6}
\end{equation}
\begin{equation}
L_1=\frac{1}{N}\sum_{i=1}^N{|}x_i-\hat{x}_i|,
\label{eq:7}
\end{equation}
\begin{equation}
L_{KL}=-\frac{1}{2} \sum_{i=1}^K\left(1+\log \left(\sigma_i^2\right)-\mu_i^2-\sigma_i^2\right),
\label{eq:8}
\end{equation}
\begin{equation}
L_{total} = L_{c} + \lambda_{\text{1}} \cdot L_{1} + \lambda_{\text{2}} \cdot L_{KL},
\label{eq:9}
\end{equation}

Here, $x \in \mathbb{R}^N$ indicates the original 128-vector feature, while $\hat{x} \in \mathbb{R}^N$ denotes the fingerprint feature vector reconstructed by the VAE. $x_i$ and $\hat{x}_i$ respectively denote the $i$-th element of the original fingerprint feature and the reconstructed fingerprint feature. Additionally, $\mu \in \mathbb{R}^K$ denotes the mean vector in the latent space, $\sigma^2 \in \mathbb{R}^K$ denotes the variance vector in the latent space, and the $KL$ divergence between the normally distributed output of the encoder $\mathcal{N}(\mu, \sigma^2)$ and the standard normal distribution $\mathcal{N}(0, 1)$.

\noindent \textbf{Model Attribution and Anonymization.}
\label{ssec:2.4.2}
Our attribution network aims to predict the source model of AI-generated images. It takes the causal fingerprint of an image obtained through VAE optimization as input and predicts the identity of its source model. The attribution network we employ utilizes a pre-trained CLIP \cite{C43} (ViT-B/32 architecture) enhanced with an attention mechanism \cite{C44} to bolster classification capabilities. It is fine-tuned using standard cross-entropy loss. The architecture diagram in Fig. \ref{fig:res2} illustrates our model's artifact extraction, causal fingerprint optimization, visual disentanglement, and attribution process.

In the counterfactual approach, we employ the Projected Gradient Descent (PGD) adversarial perturbation algorithm\cite{C52}. By applying perturbations constrained by the causal fingerprint to images, we alter model predictions while maintaining perturbations within the \(\mathcal{L}^{\infty}\) norm constraint, achieving anonymization. Its significance lies in: concealing model identity to protect privacy, enhancing attribution accuracy and robustness, isolating causal effects to reduce bias, and supporting reliable generative model development.

\begin{figure}[htb]
\centering
\includegraphics[width=0.93\linewidth]{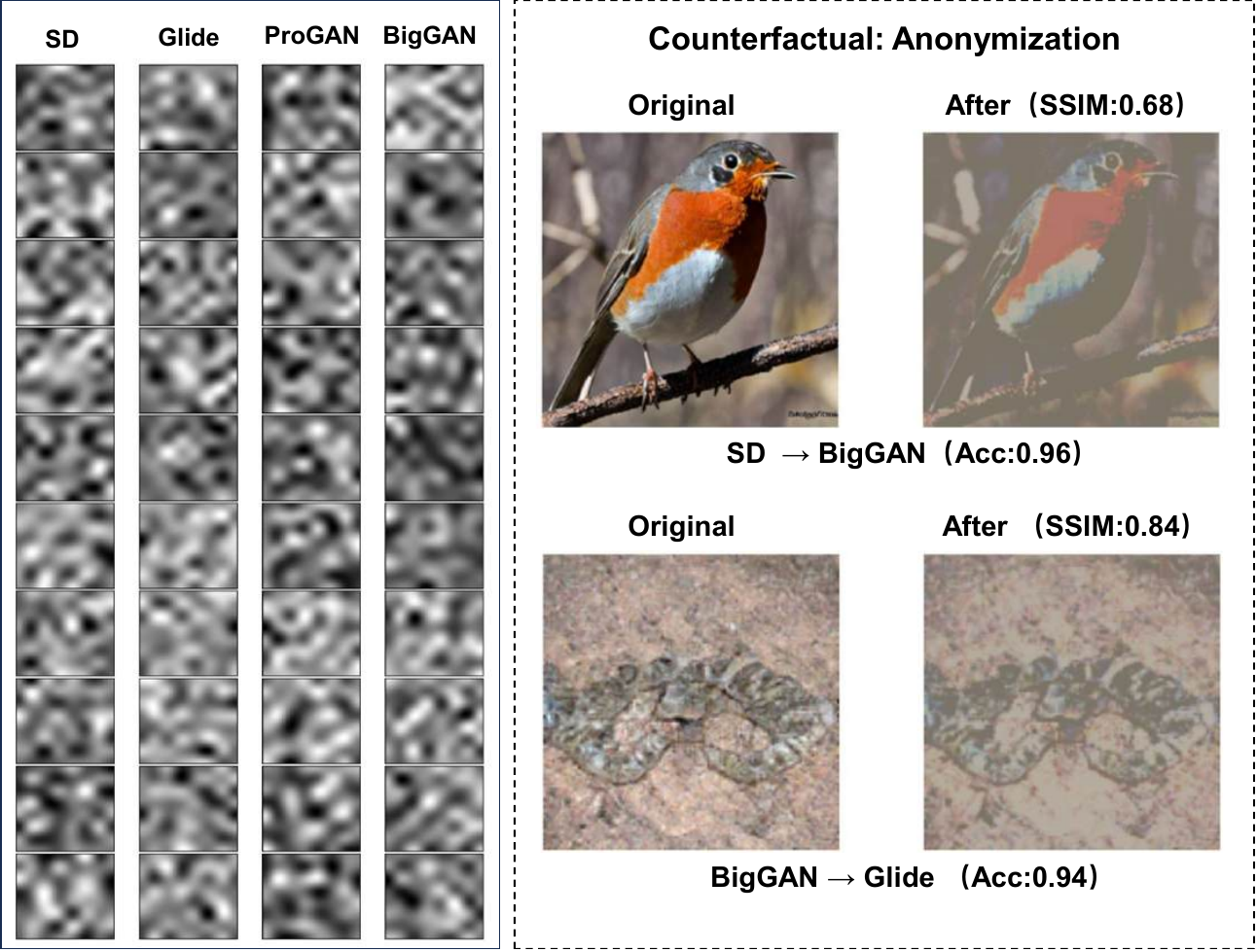}
\caption{Left: visualizations of the extracted causal fingerprints. Right: instances of successful source model anonymization via embedding counterfactual fingerprints.}
\label{fig:res4}
\end{figure}

\section{Experiments}
\label{sec:typestyle}
\subsection{Experimental Setup}
\label{ssec:subhead}

\noindent \textbf{Datasets.} To evaluate the performance of model attribution using fingerprints across diverse environments, we constructed the GM-GenImage dataset based on GenImage\cite{C26}, a large AI-generated image dataset that offers strong content and semantic diversity. We selected 10K images from each of four generative models (BigGAN, ProGAN, Glide, and Stable Diffusion (SD)), yielding a total of 40K images to construct GM-GenImage. 

\noindent \textbf{Baseline.} We evaluated four major categories of model attribution methods based on pixel-level color, frequency domain features, supervised learning, and manifold learning respectively. We compare representative methods from each group with our approach, including methods based on pixel-level colour: McCloskey et al. \cite{C22}, and Hae et al. \cite{C10}; frequency domain feature-based methods: Dzanic\cite{C23}; supervised learning-based methods: Yu et al. \cite{C9}, and Wang et al. \cite{C8}; Euclidean manifold learning-based methods: Hae et al. \cite{C11}. Their respective embedding spaces are utilized for comparative analysis.

\noindent \textbf{Settings.} We evaluated the attribution performance of the baseline methods and our method on the GM-GenImage dataset, using accuracy (\%) and the ratio of inter-class and intra-class Fréchet Distance (FDR) \cite{C48} as evaluation metrics. Attribution is achieved through extracting CFs across different embedding spaces and their ability to classify respective source models (Sec. \ref{ssec:2.2}). The classifier was trained on the training set using cross-entropy loss and evaluated on the validation and test sets for classification accuracy. Both $\lambda_{L_{1}}$ and $\lambda_{KL}$ were set to $0.1$. The Fine-gained CF obtained by expanding SILS are demonstrated in ablation study in terms of feature representations (Sec. \ref{ssec:2.3}). 

\subsection{Results}
\noindent \textbf{Model attribution.} By leveraging projection differences across multiple semantically invariant embedding spaces, we disentangled the causal fingerprint and trained a classifier to evaluate the model attribution efficacy. Correct source attribution among different models indicates that the causal fingerprint captures the source information accurately, thereby validating its existence.

Table \ref{tab:my-table1} presents the model attribution results for fingerprint recognition methods. The convolutional neural network (CNN) classifier outperforms KNN and support vector machines (SVM). Ablation experiments indicate that both the pre-trained ViT model's class-labeled features and the grayscale FFT low-frequency spatial extraction features lead to poor fingerprint attribution performance. Subsequent multi-space fusion techniques improved attribution performance by reducing weights. Additionally, the FDR metric validated the separability of fingerprint feature spaces, with higher FDR values indicating stronger attribution capability. This method achieves significantly better FDR values than competitors, with an average attribution accuracy improvement of 22.78\% (compared to 33.61\%, 16.17\%, 14\%, 15.91\%, 23.54\%, and 33.42\% for other methods), confirming the effectiveness of causal fingerprints in generative model attribution.

\begin{table}[!htp]
	\centering
        \resizebox{\linewidth}{!}{
        \setlength{\tabcolsep}{2.5mm}{
	\begin{tabular}{lcccc}
		\hline
		Method  & acc.(\%) $\uparrow$   & FDR $\uparrow$    & prec $\uparrow$  & recall $\uparrow$ \\
        \hline
		Yu et al.  (ICCV 2019) \cite{C9}            & 64.43 (+33.61) & 148.16  & 0.657 & 0.644 \\
		Wang et al.  (CVPR 2019) \cite{C8}          & 81.87 (+16.17) & 3.49   & 0.837 & 0.819 \\
		McCloskey et al. (arXiv 2018) \cite{C22}    & 84.04 (+14.00) & 3.38   & 0.843  & 0.840 \\
		Dzanic et al. (NeurIPS 2020) \cite{C23}       & 82.13 (+15.91) & 48.40  & 0.824 & 0.821 \\
        Hae et al.  (CVPR 2024) \cite{C10} & 64.62 (+33.42) & 1.24    & 0.699 & 0.646 \\ 
		Hae et al.  (arXiv 2025) \cite{C11} & 74.5 (+23.54) & 83.34  & 0.858  & 0.745  \\
		 \rowcolor{gray!40}
       \hline
		\textbf{Ours}               & 98.04 & 357.01 & 0.980 & 0.980 \\ 
        \hline
	\end{tabular}}}
	\caption{Model attribution results evaluated in the task of predicting the source generative model of generated samples.}
	\label{tab:my-table1}
\end{table}



\begin{table}[!htp]
	\centering
         \resizebox{\linewidth}{!}{
        \setlength{\tabcolsep}{5mm}{
	\begin{tabular}{ccccc}
		\hline
		Method                & acc.(\%) $\uparrow$   & FDR $\uparrow$      & prec $\uparrow$   & recall $\uparrow$ \\ \hline
		dire\_RGB              & 99.54 & 137.72 & 0.995 & 0.995 \\
		dire\_DFT              & 99.75 & 278.48 & 0.998 & 0.998 \\
		dire\_QFT              & 64.62 & 291.89 & 0.646 & 0.646 \\
		dire\_SL               & 86.25 & 196.01   & 0.862 & 0.863 \\
		dire\_SSL              & 99.67 & 216.95 & 0.997 & 0.997 \\
		dire\_VSL              & 68.17 & 128.18 & 0.684 & 0.682 \\ \hline
		multi\_fingerprint($F_{0}$) & 88.83 & 199.67  & 0.888 & 0.888 \\
		multi\_fingerprint($F_{1}$) & 94.33 & 312.77  & 0.944 & 0.943 \\
		multi\_fingerprint($F_{2}$) & 98.04 & 357.01  & 0.980 & 0.980 \\ \hline
	\end{tabular}}}
	\caption{Experimental results on artifact feature ablation. Here, $F_{0}$, $F_{1}$, and $F_{2}$ represent the unweighted, empirically weighted (proportionally weighted based on results from preceding embedding spaces), and cross-attention weighted approaches, respectively.}
	\label{tab:my-table2}
\end{table}

\noindent \textbf{Ablation study of feature representations.} 
We evaluated the artifact characteristics of different semantically invariant embedding spaces in the source model task for predicting generated samples. As shown in Table \ref{tab:my-table2}, the highest attribution accuracy exceeding 99\% was achieved in the RGB, DFT, and SSL spaces, while accuracy was slightly lower in the VSL and QFT spaces. This discrepancy may stem from the attribution network learning self-supervised features and frequency-domain features more effectively than supervised features and low-frequency components. Regarding the final multi-space fusion, both the fingerprint attribution accuracy without weighting ($F_{0}$) and the empirically weighted approach ($F_{1}$) based on prior attribution accuracy results outperformed the weighting obtained through cross-attention learning ($F_{2}$). These findings provide a basis for subsequent attribution experiments on causal fingerprints.

\noindent \textbf{Visualization and source anonymization.} Fig. \ref{fig:res4} displays the visual causal fingerprint structures extracted from four models, accompanied by causal relationship examples. By anonymizing the causal fingerprints of source-generation models using a counterfactual approach, we demonstrate that causal fingerprints maximize only their own model responses and remain unaffected by non-causal representations. With recognition accuracy consistently exceeding 90\%, this supports the validity of the attribution mechanism and validates the existence of causal fingerprints.




\section{Conclusion}
\label{sec:majhead}

From a causal inference perspective, we investigate solutions to the attribution challenge in image source generation models. By focusing on underlying causal relationships, we propose a formalized causal decoupling method and define causal fingerprints, filling a gap in model forensics research. Experiments validate the significant advantages of causal fingerprints in distinguishing AIGC models such as BigGAN, ProGAN, Glide, and Stable Diffusion, with attribution performance surpassing existing methods. This approach enhances the safety of AIGC content and lays the foundation for multimedia visual forensics research.

\section{Acknowledgment}
This work was supported by the National Natural Science Foundation of China under Grant No. 62402009, the Key Laboratory of Computing Power Network and Information Security, Ministry of Education under Grant No.2024PY014, and the Science and Technology Development Fund of Macao under Grant No. 0013-2024-ITP1 and 0069/2025/ITP2.

\footnotesize
\bibliographystyle{IEEEbib}
\bibliography{strings,refs}

\end{document}